%% file: main.tex
\documentclass[10pt,twocolumn,letterpaper]{article}

\usepackage[pagenumbers]{cvpr} 
\usepackage{multirow}
\usepackage{booktabs}
\usepackage[ruled]{algorithm2e} 
\usepackage{xcolor}  
\usepackage{amsmath}  
\usepackage{amssymb}  

\newcommand{\parsection}[1]


\definecolor{cvprblue}{rgb}{0.21,0.49,0.74}
\usepackage[pagebackref,breaklinks,colorlinks,allcolors=cvprblue]{hyperref}

\def\paperID{7307} 
\def\confName{CVPR}
\def\confYear{2025}

\title{GradiSeg: Gradient-Guided Gaussian Segmentation with Enhanced 3D Boundary Precision}

\author{\textbf{Zehao Li$^{1,2}$}, \textbf{Wenwei Han$^{1,2}$}, \textbf
{Yujun Cai$^{3}$}, \textbf{Hao Jiang$^{1,2*}$}\\ \textbf{Baolong Bi$^{1,2}$}, \textbf{Shuqin Gao$^{1}$}, \textbf{Honglong Zhao$^{1}$}, \textbf{Zhaoqi Wang$^{1,2}$}\\
\vspace{-0.3cm} 
\\
$^1$Institute of Computing Technology, Chinese Academy of Sciences, ICT \\
$^2$University of Chinese Academy of Sciences, UCAS \\
$^3$The University of Queensland\\
{\tt\small \{lizehao23z, hanwenwei23s\}@ict.ac.cn, yujun.cai@uq.edu.au}\\
{\tt\small\{jianghao, bibaolong23z, gaoshuqin, zhaohonglong, zqwang\}@ict.ac.cn}
}

\begin{document}

\maketitle

\input{sec/0_abstract}    
\input{sec/1_intro}
\input{sec/2_rel}
\input{sec/3_met}
\input{sec/4_exp}
\input{sec/5_con}
{
    \small
    \bibliographystyle{ieeenat_fullname}
    \bibliography{main}
}
\input{sec/X_suppl}

\end{document}

%% file: sec/0_abstract.tex
\begin{abstract}
While 3D Gaussian Splatting enables high-quality real-time rendering, existing Gaussian-based frameworks for 3D semantic segmentation still face significant challenges in boundary recognition accuracy. To address this, we propose a novel 3DGS-based framework named GradiSeg, incorporating Identity Encoding to construct a deeper semantic understanding of scenes. Our approach introduces two key modules:  Identity Gradient Guided Densification (IGD) and Local Adaptive K-Nearest Neighbors (LA-KNN). The IGD module supervises gradients of Identity Encoding to refine Gaussian distributions along object boundaries, aligning them closely with boundary contours. Meanwhile, the LA-KNN module employs position gradients to adaptively establish locality-aware propagation of Identity Encodings, preventing irregular Gaussian spreads near boundaries. We validate the effectiveness of our method through comprehensive experiments. Results show that GradiSeg effectively addresses boundary-related issues, significantly improving segmentation accuracy without compromising scene reconstruction quality. Furthermore, our method's robust segmentation capability and decoupled Identity Encoding representation make it highly suitable for various downstream scene editing tasks, including 3D object removal, swapping and so on.
\end{abstract}

%% file: sec/1_intro.tex
\section{Introduction}
\label{sec:intro}

3D semantic segmentation aims to assign semantic labels to different objects and regions in a 3D scene, providing a comprehensive understanding of the scene. It serves as a fundamental task for various applications, such as autonomous driving, robotic manipulation, and virtual reality~\citep{Wang_2024_CVPR,Kong_2023_ICCV,schieber2024semanticscontrolledgaussiansplattingoutdoor, yang2023unified}. However, achieving precise and detailed 3D semantic segmentation remains challenging due to the inherent complexity and diversity of real-world scenes, particularly around intricate object boundaries.

\begin{figure}[t]
  \centering
  \includegraphics[width=1.0\linewidth]{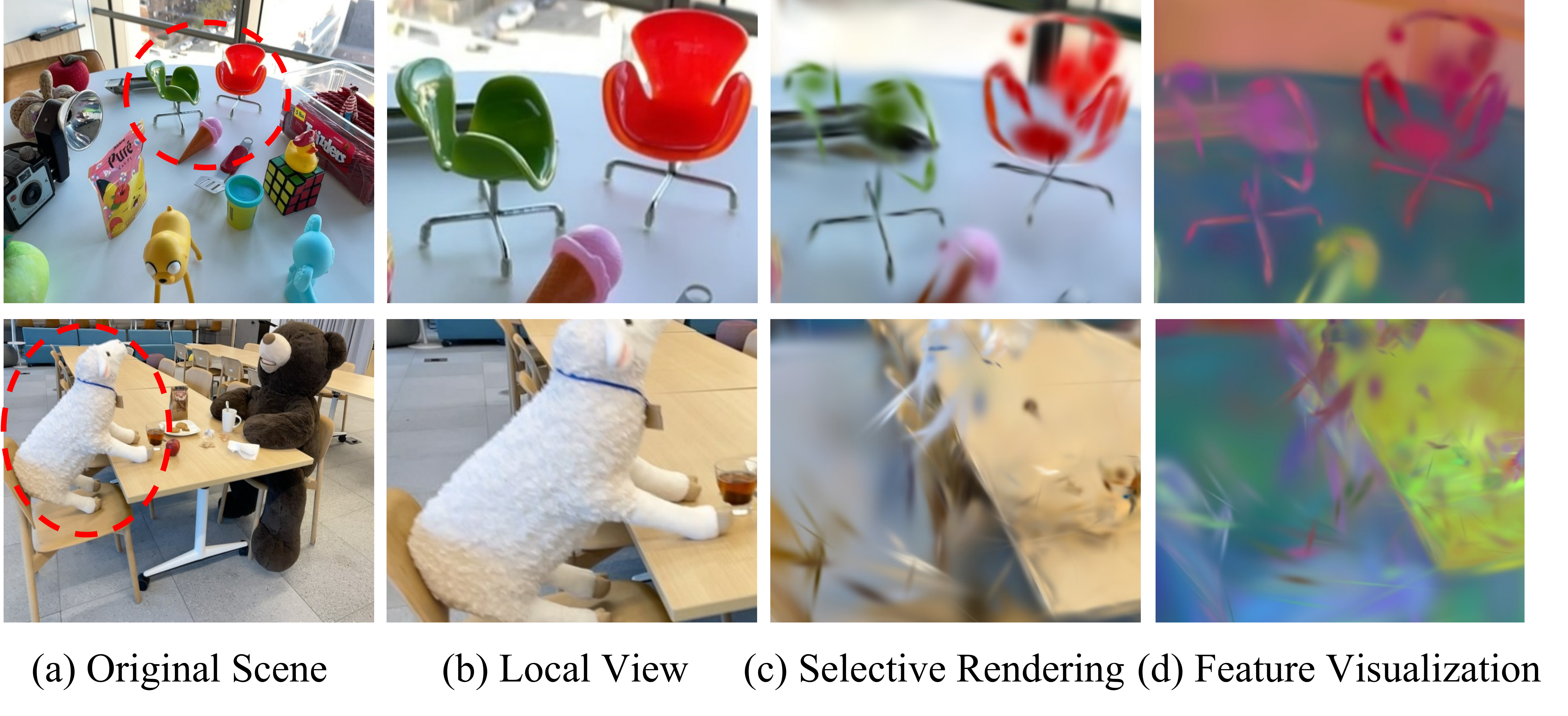}
  \caption{We adopt Identity Encoding to construct 3D semantic segmentation. In the original 3D scene (column a), we selectively render Gaussians that exhibit unusually high Identity Encoding gradients, generating a selective rendering (column c). It is observed that these Gaussians predominantly cluster around object boundaries. To facilitate comparison, we present a locally enlarged view of the original 3D scene (column b) and visualize the Identity Encoding features of these selected Gaussians (column d).}
   \label{fig:visual}
\end{figure}
Although substantial progress has been achieved in 2D semantic segmentation, especially with foundation models like the Segment Anything Model (SAM)~\citep{kirillov2023segment}, which generalize effectively to unseen objects and scenes, extending these advancements to 3D remains a significant challenge. Unlike 2D data, 3D data involves more complex spatial structures and higher dimensionality, making it difficult to achieve generalization in a same level. Additionally, the limited availability of large-scale annotated 3D datasets exacerbates the challenges in 3D semantic segmentation.

In response to the unique challenges of 3D segmentation, recent approaches have shifted towards advanced 3D scene representations such as Neural Radiance Fields (NeRF)~\citep{mildenhall2021nerf} and 3D Gaussian Splatting (3DGS)~\citep{kerbl20233d}. NeRF-based methods~\citep{cen2023segment, zarzar2022segnerf, ying2024omniseg3d, zhi2021place, yang2021tupper} primarily incorporate 2D semantic information into neural radiance fields for rendering. However, the dense sampling strategy used in NeRF is computationally expensive, making it difficult to achieve real-time performance. On the other hand, 3DGS-based methods~\citep{cen2024segment3dgaussians, ye2023gaussian} leverage the explicit representation of a scene with a set of Gaussians, resulting in achieving high-quality reconstruction and real-time rendering. Nevertheless, current 3DGS-based methods still suffer from blurry and imprecise object boundaries in the segmentation results. The coarse modeling of Gaussian distributions fails to accurately capture the subtle features at object boundaries, resulting in suboptimal segmentation accuracy.

Given the persistent segmentation ambiguities at object boundaries observed in existing methods, we propose \textbf{Gradi}ent-Guided Gaussian \textbf{Seg}mentation (\textbf{GradiSeg}) to enhance 3D Boundary Precision, which adopts Identity Encoding to construct the model's semantic understanding of 3D scenes. Notably, GradiSeg leverages gradient information to adaptively refine Gaussian distributions near object boundaries, improving the representation of Identity Encoding features in local space and enabling more precise boundary-aware semantic segmentation.

Specifically, we introduce two key modules: Identity Gradient Guided Densification (IGD) and Local Adaptive K-Nearest Neighbors (LA-KNN). Gaussians situated at or near object boundaries tend to exhibit higher gradients in their Identity Encoding during training, adapting to capture the semantic transitions across these boundaries, as shown in Figure~\ref{fig:visual}. IGD detects and utilizes these gradient signals to adaptively split and adjust the Gaussians at object boundaries to refine the Gaussian distributions, resulting in more accurate boundary modeling. Meanwhile, LA-KNN identifies the adaptive local neighborhood for each Gaussian based on the gradients of Gaussians' positions and enforces the consistency of their Identity Encodings. This local consistency constraint improves the continuity and accuracy of boundary segmentation and prevents irregular Gaussian spreads near boundaries, thereby promoting stable convergence of the scene.

We validate the effectiveness of GradiSeg through comprehensive experiments on two public 3D scene datasets: LERF-Mask~\citep{ye2023gaussian} and Mip-NeRF 360~\citep{barron2022mip}. The results show that GradiSeg significantly outperforms state-of-the-art methods in 3D semantic segmentation without compromising reconstruction quality, despite modifying the Gaussian distribution of the scene. Compared to the baseline Gaussian Grouping method, GradiSeg achieves an average improvement of 5.27\% in mean Intersection over Union (mIoU) on the LERF-Mask dataset, with a maximum improvement of 11.6\%. Additionally, it demonstrates a 6.3\% average improvement in mean Boundary Intersection over Union (mBIoU), with a maximum improvement of 10.2\%. Furthermore, we demonstrate GradiSeg's broad applicability in supporting various downstream applications such as interactive scene removal and swapping, benefiting from its precise boundary modeling.

In summary, our main contributions are as follows:
\begin{itemize}
    \item  We propose Gradient-Guided Gaussian Segmentation (GradiSeg), a novel 3DGS-based framework for 3D semantic scene segmentation that effectively alleviates the challenge of blurry and imprecise object boundaries.
    \item  Our framework integrates two innovative modules: IGD, which refines boundary Gaussians by adaptively detecting, segmenting, and adjusting them, and LA-KNN, which enforces local consistency in Identity Encodings. Together, these modules enable more precise semantic modeling at boundaries, significantly reducing boundary blurriness and enhancing overall segmentation quality.
    \item   We conduct comprehensive experiments on both LERF-Mask~\citep{ye2023gaussian} and Mip-NeRF 360~\citep{barron2022mip}, demonstrating the superior performance of GradiSeg in 3D semantic segmentation tasks, achieving a state of the art without compromising reconstruction quality.
\end{itemize}

%% file: sec/2_rel.tex
\section{Related Work}
\label{sec:rel}

\subsection{Scene Representation for 3D Reconstruction}
\begin{figure*}[t]
  \centering
  \includegraphics[width=1.0\linewidth]{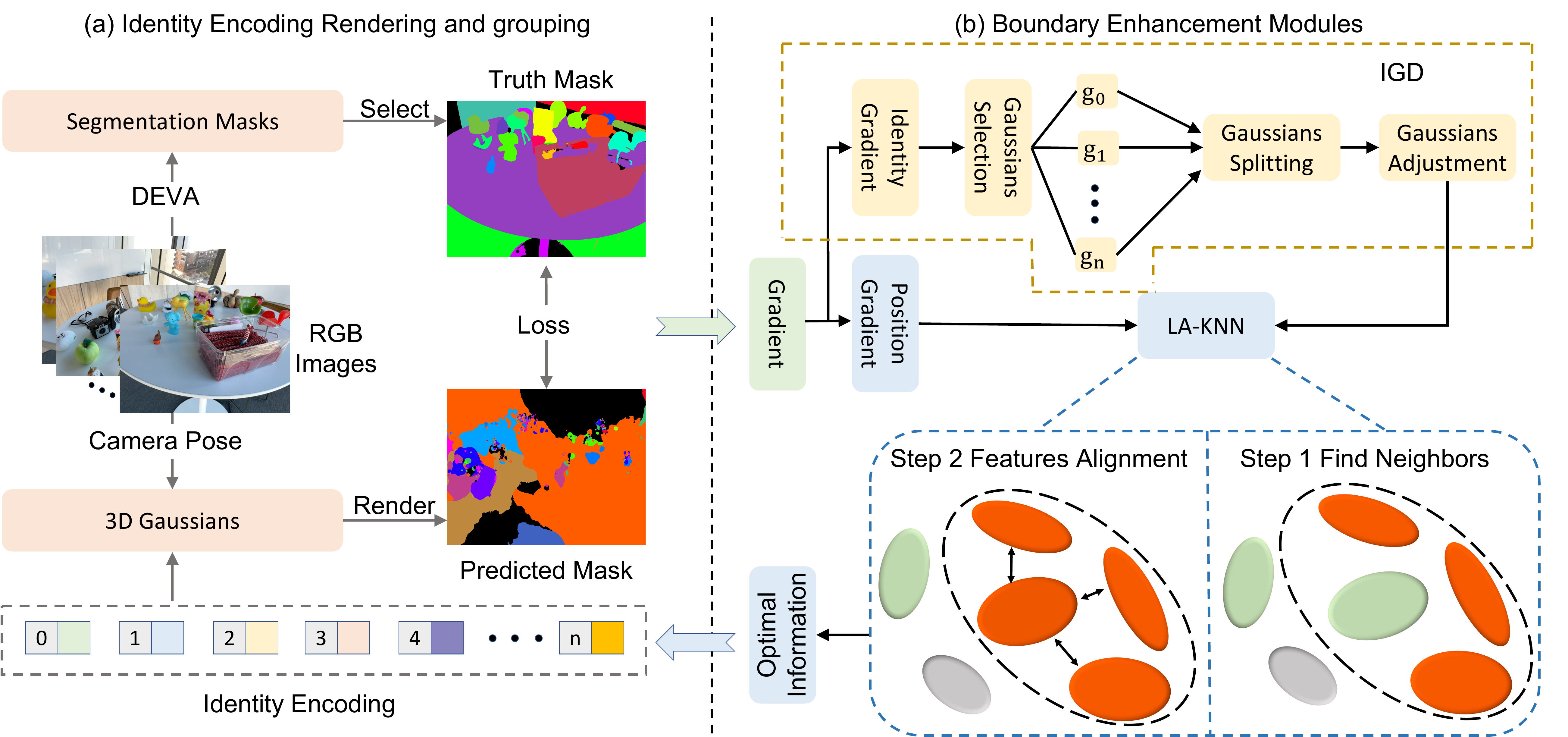}
  \caption{Overview of the proposed method. a) We adopt Identity Encoding as a learnable vector to construct a semantic understanding of the scene. This vector is optimized through multi-view supervision to produce initial segmentation results. b) To tackle boundary ambiguity, we introduce two boundary enhancement modules: IGD and LA-KNN. IGD refines Gaussians near object boundaries by monitoring Identity Encoding gradients. Complementarily, LA-KNN enables direction-aware feature propagation by leveraging position gradients for neighbor selection, preventing cross-instance feature contamination at boundaries.}
   \label{fig:pipeline}
\end{figure*}

In recent years, numerous outstanding works on 3D scene representation have rapidly emerged. Among them, NeRF~\citep{mildenhall2021nerf}, which uses implicit neural radiance fields to represent 3D scenes, enables high-quality scene rendering. Several methods~\citep{zhang2020nerf++, barron2022mip, liu2020neural, fridovich2022plenoxels, muller2022instant} have introduced improvements to NeRF, significantly enhancing both the quality and efficiency of scene reconstruction. However, the dense sampling strategy of NeRF requires substantial computational resources, leading to slow rendering speeds. Compared to implicit representation methods, 3D Gaussian Splatting (3DGS)~\citep{kerbl20233d} has been proposed as an explicit way to represent 3D scenes using a set of 3D Gaussians, enabling high-quality, real-time scene rendering, but the discrete nature of the Gaussian representation may lead to a slight decrease in accuracy. A series of works in various application domains within the 3DGS framework have gradually emerged. For instance, several strategies have been proposed for dynamic scene reconstruction~\citep{luiten2023dynamic, yang2024deformable, wu20244d,lin2023gaussianflow4dreconstructiondynamic}. Some methods have leveraged the explicit representation of 3DGS to achieve quality enhancement~\citep{yu2023mipsplattingaliasfree3dgaussian,yan2024multiscale3dgaussiansplatting,lu2023scaffoldgsstructured3dgaussians}, acceleration~\citep{girish2024eaglesefficientaccelerated3d,hamdi2024gesgeneralizedexponentialsplatting,lin2024rtgsenablingrealtimegaussian,hu2024lowlatencypointcloud} and large scene reconstruction~\citep{zhang2024garfieldreinforcedgaussianradiance,liu2024citygaussianrealtimehighqualitylargescale,cheng2024gaussianpro3dgaussiansplatting,lin2024vastgaussianvast3dgaussians}.

\subsection{3D Semantic Segmentation}
The rapid progress in 3D scene reconstruction has provided a solid foundation for advancements in 3D semantic segmentation. The majority of these methods are predominantly grounded in NeRF and 3DGS frameworks.
\paragraph{NeRF-based 3D Semantic Segmentation}  Initially, SemanticNeRF~\citep{zhi2021place} achieves a joint implicit representation of geometry, appearance, and semantics by injecting semantic information into NeRF, enabling both semantic labeling and semantic view synthesis. Subsequently, by providing only manual segmentation prompts to generate a 2D mask in a given view using SAM~\citep{kirillov2023segment}, SA3D~\citep{cen2023segment} can iteratively construct the 3D mask of the target object within a voxel grid. Furthermore, various approaches~\citep{ying2024omniseg3d,kim2024garfield} have been introduced to perform the segmentation of objects within a scene at different granularities, enabling fine-tuned differentiation across multiple scales. Although the aforementioned methods enable 3D semantic segmentation based on NeRF, the excessive sampling strategies result in high computational costs and hinder real-time rendering.
\paragraph{3DGS-based 3D Semantic Segmentation} 
Leveraging the real-time rendering capability of 3DGS, several semantic segmentation methods built upon this framework have emerged. SAGA~\citep{cen2024segment3dgaussians} combines SAM with 3DGS and, through well-designed contrastive training, effectively embeds the 2D segmentation results generated by the segmentation model into 3D Gaussian point features. Similarly, FeatureGS~\citep{zhou2024feature} and Click-GS~\citep{choi2024click} extract the feature fields of 2D foundational models into 3DGS, enabling tasks such as novel view semantic segmentation and segmentation of arbitrary objects. Unlike the aforementioned methods that embed 2D features into 3D Gaussians, Gaussian Grouping~\citep{ye2023gaussian} introduces new attributes directly into the Gaussians. This approach integrates them directly into the Gaussian rendering process to accomplish segmentation tasks, facilitating the handling of subsequent downstream tasks. Different from Gaussian Grouping which solely introduces additional parameters into Gaussians, our method specifically focuses on addressing boundary ambiguities through gradient-guided optimization, leading to more precise segmentation at object boundaries.

%% file: sec/3_met.tex
\section{Methodology}
\label{sec:met}
\subsection{Overview}
Our goal is to achieve precise 3D semantic segmentation, with a focus on boundary regions where existing methods often produce ambiguous results. To address this, we propose GradiSeg, a gradient-guided framework that enhances boundary segmentation through adaptive Gaussian refinement. As shown in Figure~\ref{fig:pipeline}, GradiSeg combines Identity Encoding to establish basic scene semantics, where each Gaussian holds a learnable encoding vector for semantic grouping, optimized through multi-view supervision to produce initial segmentation results. 

To tackle boundary ambiguity, we introduce two novel modules: Identity Gradient Guided Densification (IGD) and Local Adaptive K-Nearest Neighbors (LA-KNN). The IGD module identifies and refines Gaussians near object boundaries by monitoring gradients of Identity Encodings. Complementing IGD, the LA-KNN module achieves direction-aware local feature propagation. Together, these modules work synergistically to enhance boundary precision while maintaining semantic consistency.

The framework is trained end-to-end with a joint objective that integrates reconstruction accuracy, semantic segmentation, and boundary semantic enhancement. In the following sections, we briefly review the 3D Gaussian Splatting in Sec \ref{sec:pre}. We then illustrate how Identity Encoding is utilized for rendering and grouping in Sec \ref{subsec:id}. After that, we detail the IGD and LA-KNN in Sec \ref{subsec:iegd} and Sec \ref{subsec:la-knn}.

\subsection{Preliminary: 3D Gaussian Splatting}
\label{sec:pre}
\begin{figure}[t]
  \centering
  \includegraphics[width=1\linewidth]{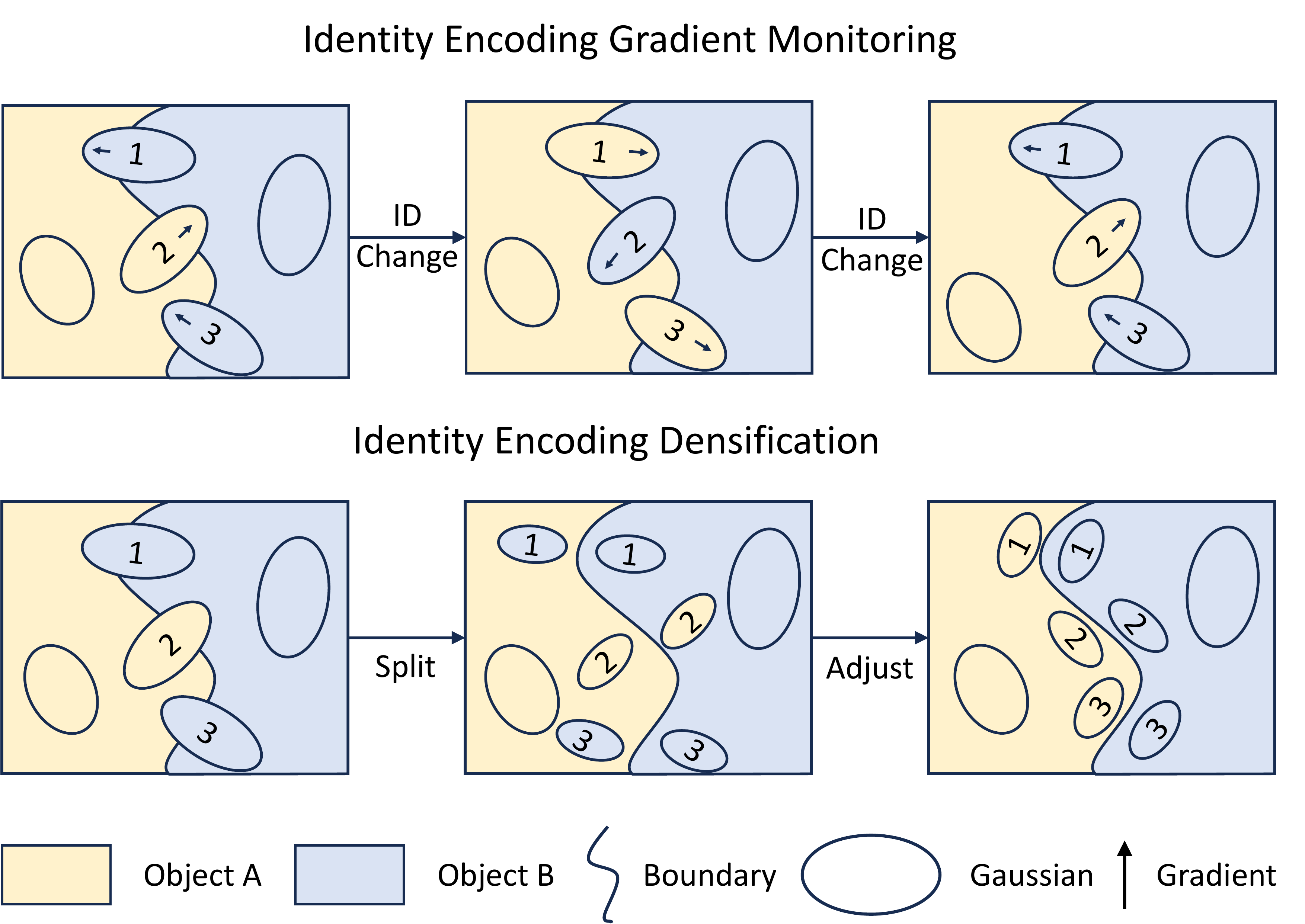}
  \caption{The process of the IGD module. The first row refers to Identity Encoding gradient monitoring. For Gaussians near the boundaries, in order to optimize, they continuously adjust their Identity Encoding, leading to an increasingly high gradient that may become anomalous. The second row involves Identity Encoding densification. For Gaussians with anomalous gradients, we perform splitting and adjust them to both sides of the boundary, addressing optimization conflicts during the training process.}
   \label{fig:iegd}
\end{figure}

3D Gaussian Splatting (3DGS) can explicitly represent a 3D scene and achieve high-quality rendering with real-time efficiency. Beginning with a set of pose-annotated images $I=\{I_1,I_2,...,I_V\}$ captured from multiple viewpoints, we initialize and construct learnable 3D Gaussians $G=\{g_1,g_2,...,g_N\}$, where V denotes the numbers of input images, and N denotes the number of 3D Gaussians in the 3D scene. Each 3D Gaussian $g_i$ is composed of multiple learnable parameters $\{p_i,s_i,r_i,o_i,c_i\}$. Furthermore, $p_i=\{x_i,y_,z_i\}\in\Bbb{R}^3$ is the position of $g_i$ in 3D space. $s_i\in\Bbb{R}^3$ is the scaling factor and $r_i\in\Bbb{R}^4$ is the rotation quaternion of $g_i$, they are used to represent the 3D covariance matrix. $o_i\in\Bbb{R}$ denotes the opacity, and $c_i$ represents the color in the form of spherical harmonics (SH) coefficients. Given a camera pose, the 3D Gaussians are projected onto a 2D plane, enabling fast differentiable rasterization. The color 
C of a pixel is computed through $\alpha$-blending with depth-ordered N Gaussians overlapped the pixel:
\begin{equation}
C = \sum_{i \in \mathcal{N}} c_i \alpha_i \prod_{j=1}^{i-1} (1 - \alpha_j),  
\end{equation}
where $\alpha_i$ is the blending weight computed by $o_i$, the proje cted 2D covariance of $g_i$ and pixel distance.

\subsection{Identity Encoding Rendering and Grouping}
\label{subsec:id}

To establish a comprehensive semantic understanding of the 3D scene, following~\citep{ye2023gaussian}, we employ DEVA~\citep{cheng2023trackingdecoupledvideosegmentation} to produce multi-view consistent segmentation masks and introduce Identity Encoding into the Gaussian. This attribute, represented as a vector, is designed to differentiate between instance groups in 3D space. It functions similarly to the color attribute, which is also represented using spherical harmonics but limited to the zeroth order, ensuring group consistency across multiple viewpoints. The Identity Encoding rendering is similar to color rendering, as it is also projected onto the pixel plane for $\alpha$-blending operations:
\begin{equation}
E_{id} = \sum_{i \in \mathcal{N}} e_i \alpha_i \prod_{j=1}^{i-1} (1 - \alpha_j), 
\end{equation}
where $E_{id}$ is the 2D Identity Encoding of the pixel, $e_i$ is the i-th depth-ordered Gaussian's Identity Encoding and $\alpha_i$ is the blending weight. $E_{id}$ is computed as the weighted average of 3D Identity Encoding, making it a vector like $e_i$.

After obtaining the 2D Identity Encoding for all pixels, a classification neural network is used to output the probability of each pixel belonging to different groups, thereby using cross-entropy loss as $L_{2d}$. Upon completion of model training, the 3D Gaussians corresponding to each instance group can be identified based on the Identity Encodings. This enables various downstream tasks, such as group deletion, group rendering and group style transfer.

\subsection{Identity Gradient Guided Densification}
\label{subsec:iegd}

We observe that boundary conflicts can arise during Identity Encoding Rendering, as Gaussians located near group boundaries are assigned a single Identity Encoding, as shown in Figure~\ref{fig:visual}. This assignment leads to optimization conflicts at the boundaries during loss minimization, resulting in instability in the learning process of Identity Encodings. Gaussians at the boundaries continually adjust their Identity Encodings to accommodate this conflicting situation, which causes the gradients of Identity Encodings to accumulate without alleviating the boundary segmentation conflict, as shown in Figure~\ref{fig:iegd}. 

To address this issue, we monitor the accumulated gradients of the Identity Encoding to identify Gaussians positioned near object boundaries. When the accumulated gradient exceeds a specified threshold, we perform  Identity Gradient Guided Densification to refine the Gaussian distribution, ensuring that Gaussians are accurately positioned on both sides of the object boundary.
\begin{algorithm}[t]
    \scriptsize 
    \caption{\textit{Identity Gradient Guided Densification (IGD)}}
    \label{alg:IEGD}
    \KwIn{The Gaussians $G=\{g_1,g_2,...,g_N\}$ generated in each epoch during the model training process. The $\nabla L$ represents the gradient information propagated back during the training process. The current training iteration $iter$.}
    \KwOut{The Gaussians $G^{\prime}$ refined by the IGD}
    \If{\textnormal{IsIGDIteration}($iter$)}{
        $p, s, o, c, e, gradient \gets \text{GetAttributes}(G)$\\
        \textcolor{blue}{// Positions, Covariances, Opacities, Colors, Identity Encodings and Accumulated Gradient of Identity Encodings.}
        
        \ForEach{$g_i$ in $G$}{
            $gradient_i \gets gradient_i + \nabla_{e_i} L$\\
            \textcolor{blue}{// Identity Encoding Gradient Monitoring.}
            
            \If{$o_i < \epsilon$ \textbf{or} \textnormal{IsTooLarge}($p_i, s_i$)}{
                \textcolor{blue}{// Pruning.}\\
                RemoveGaussian($g_i$)
            }
            
            \If{$gradient_i > \tau$}{
                \textcolor{blue}{// Optimization of Gaussian Distribution.}
                SplitGaussian($p_i, s_i, o_i, c_i, e_i$)
                $gradient_i \gets 0$\\
                \textcolor{blue}{// Gradient Resetting.}\\
                AdjustGaussian($p_i, s_i$)
            }
        }
    }
    $G^{\prime} \gets G$\\
    \Return Gaussians $G^{\prime}$
\end{algorithm}
Specifically, in the densification process, we perform a splitting operation on Gaussians located near boundaries, dividing each into two sub-Gaussians positioned on either side of the segmentation boundary and adjusting their position and scale to align them closely with boundary contours, as shown in Figure~\ref{fig:iegd}. This allows the Identity Encodings of the two resulting sub-Gaussians to be learned in different optimization directions, avoiding the conflict that arises when a single Gaussian’s identity encoding needs to be optimized in opposite directions. Details of our algorithm are presented in Algorithm~\ref{alg:IEGD}. Essentially, this approach can assign multiple Identity Encodings to the original Gaussians, allowing for adaptive dynamic grouping based on these encodings without significantly increasing computational resource consumption.

Notably, the IGD is deferred until after the initial Gaussian densification process is complete, rather than being applied at the onset of training. This approach is predicated on the assumption that the model's semantic understanding of the scene should be grounded in a sufficiently developed geometric and appearance representation, encompassing comprehensive information such as color, to facilitate a more stable and efficient convergence of the identity encoding.

 \begin{figure}[t]
  \centering
  \includegraphics[width=1\linewidth]{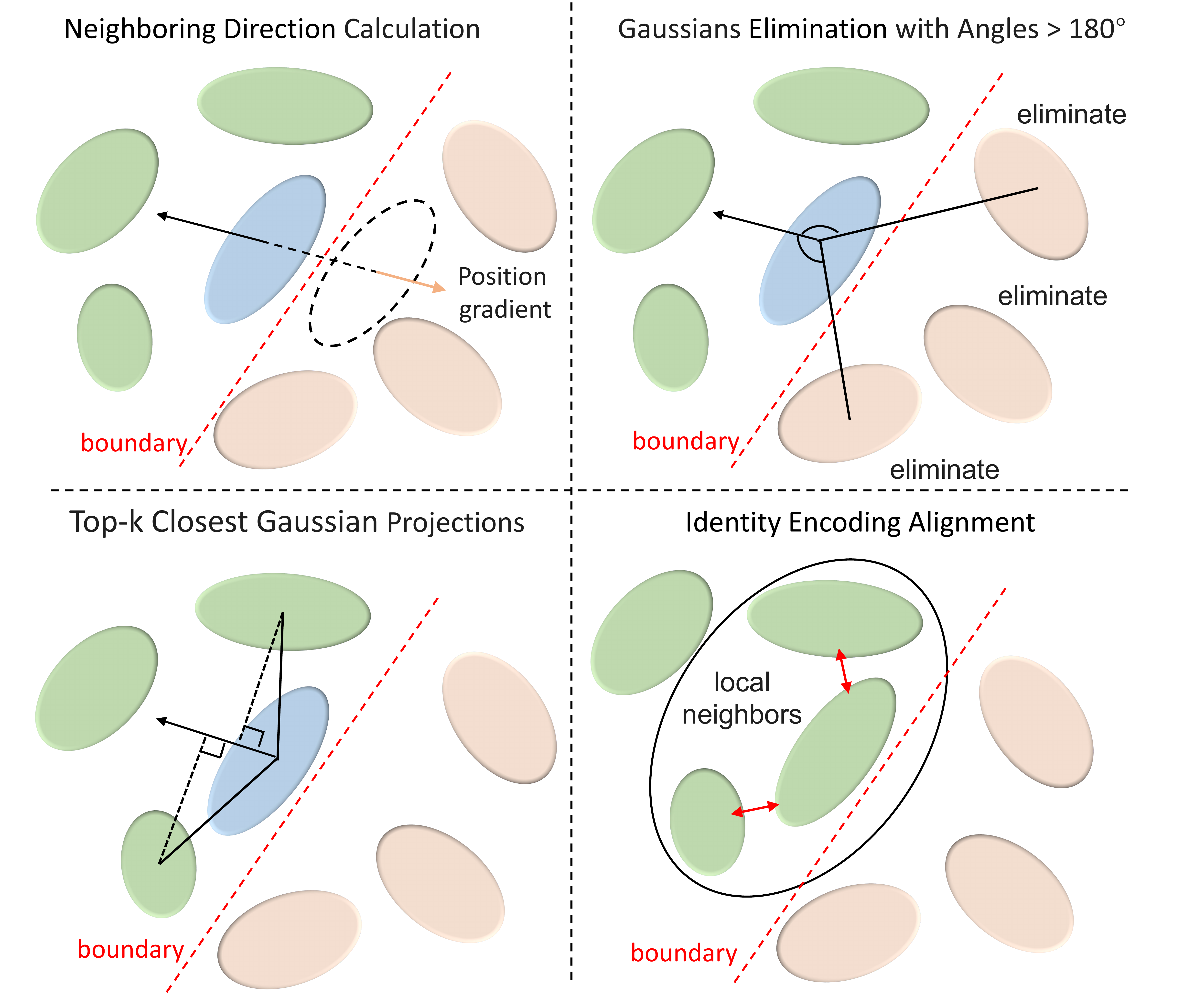}
  \caption{The process of the LA-KNN module. We first compute the neighboring direction by taking the opposite direction of the Gaussian position gradient. Then, we eliminate all Gaussians whose angle with the direction vector is greater than 180 degrees. For the remaining Gaussians, we sort them by their projection distance to the direction vector and select the $K$ nearest neighbors, where $K=2$. Finally, we align the Identity Encoding features in the local space.}
   \label{fig:la-knn}
\end{figure}

\subsection{Local Adaptive K-Nearest Neighbors}
\label{subsec:la-knn}
A key issue in the rendering process of Identity Encoding is the lack of supervision in 3D space, which limits the effective propagation of Identity Encoding throughout the 3D space. To further enhance model training and establish supervision for Identity Encoding in 3D space, we introduce the LA-KNN module. 

We sample all Gaussians in 3D space and identify the $k$ nearest local adaptive neighbors for each Gaussian, rather than searching for global nearest neighbors. Subsequently, we employ a KL divergence loss to encourage the compactness of the Identity Encodings of Gaussians within the local neighborhood. The formal representation is as follows:
\begin{equation}
L_{3d}= \frac{1}{MK}\sum_{i \in M} \sum_{j \in K}F(e_i) \log(\frac{F(e_i)}{F(e_j)})
\end{equation}
where M denotes the number of sampled Gaussians, K denotes the number of local adaptive neighbor Gaussians, and F represents the operation of feeding the Identity Encoding of each Gaussian into the neural network for classification.

In selecting local adaptive neighbors, we prioritize Gaussians that are closest to the target Gaussian along its neighboring direction, rather than simply choosing Gaussians based on euclidean distance in space as one would when searching for global nearest neighbors, as shown in Figure~\ref{fig:la-knn}. Specifically, we define the neighboring direction of the target Gaussian as the opposite of the gradient direction of its center position during training. We then compute the projection distances of all other Gaussians relative to the target Gaussian along this neighboring direction's unit vector. The $K$ Gaussians with the smallest positive projection distances are selected as the local adaptive neighbors of the target Gaussian:
\begin{equation}
\begin{aligned}
&S_k = \{ \text{Top-k}(d_i \mid d_i > 0) \}\\
&d_i = (p^{\prime} - p) \cdot u
\end{aligned}
\end{equation}
where $\text{Top-k}$ refers to the k smallest values, and ${d}_i$ represents the projection distance of the i-th Gaussian relative to the target Gaussian along the neighboring direction. Furthermore, $p$ denotes the position of the target Gaussian, $p^{\prime}$represents the position of the another Gaussian, and $u$ is the unit direction vector corresponding to the neighboring direction of the target Gaussian.

For the target Gaussian, the nearest global K Gaussians in terms of distance may not necessarily belong to the same instance, especially for Gaussians near object boundaries. Therefore, using local adaptive neighbors based on the neighboring direction vector effectively addresses this issue, preventing optimization conflicts in Identity Encoding during training and achieving direction-aware local feature propagation.

With the introduction of LA-KNN, our loss function incorporates supervision over the Gaussian Identity Encoding in 3D space. The overall loss function is formally expressed as:
\begin{equation}
L= L_1(I_{in},I_{out})+{\alpha}L_{2d}+{\beta}L_{3d}
\end{equation}
where $I_{in}$ is the input RGB image, $I_{output}$ is the is the rendered RGB image, $L_{2d}$ is the 2D Loss, $L_{3d}$ is the 3D Loss.

%% file: sec/4_exp.tex
\section{Experiments}
\label{sec:exp}
\begin{table}[t]
\centering
\resizebox{1\linewidth}{!}{
\begin{tabular}{l|cc|cc|cc}
\toprule
\multirow{2}{*}{Method} & \multicolumn{2}{c|}{figurines} & \multicolumn{2}{c|}{ramen} & \multicolumn{2}{c}{teatime}  \\ 
& mIoU & mBIoU & mIoU & mBIoU & mIoU & mBIoU    \\ \midrule 
DEVA~\citep{cheng2023trackingdecoupledvideosegmentation}&46.2 & 45.1 & 56.8  & 51.1 & 54.3 & 52.2 \\
LERF~\citep{kerr2023lerflanguageembeddedradiance} & 33.5 & 30.6 & 28.3 & 14.7 & 49.7 & 42.6  \\
SA3D~\citep{cen2023segment} & 24.9 & 23.8 & 7.4 & 7.0 & 42.5 & 39.2\\
LangSplat~\citep{qin2024langsplat3dlanguagegaussian} & 52.8 & 50.5 & 50.4 & 44.7 & 69.5 & 65.6\\
Gau-Group~\citep{ye2023gaussian} & 69.7 & 67.9 & 77.0 & 68.7 & 71.7 & 66.1  \\
GradiSeg (ours) & \textbf{81.3} & \textbf{78.1} & \textbf{78.5} & \textbf{72.9} & \textbf{74.4} & \textbf{70.6} \\ 
\bottomrule
\end{tabular}}
\caption{Comparison of open vocabulary segmentation on LERF-Mask dataset~\citep{ye2023gaussian}, focusing on three scenes: figurines, ramen, and teatime. We adopt the detections from Grounding DINO~\citep{liu2023grounding} to select mask IDs in a 3D scene like Gaussian Grouping~\citep{ye2023gaussian}.}
\vspace{-2mm}
\label{tab:lerf}
\end{table}
\subsection{Dataset and Experiment Setup}
\paragraph{Datasets}
To evaluate the semantic segmentation performance of our method, we conduct experiments on the LERF-Mask dataset~\citep{ye2023gaussian} across open vocabulary and multi-view semantic segmentation tasks. The dataset comprises three detailed scenes. Each is annotated with multiple text queries and corresponding precise mask annotations. To evaluate reconstruction quality, we test our method on the 9 scene sets provided in Mip-NeRF 360~\citep{barron2022mip}. To reduce memory consumption, we downsample the dataset by a factor of 8 prior to conducting evaluations.
\begin{figure*}[t]
  \centering
  \includegraphics[width=1\linewidth]{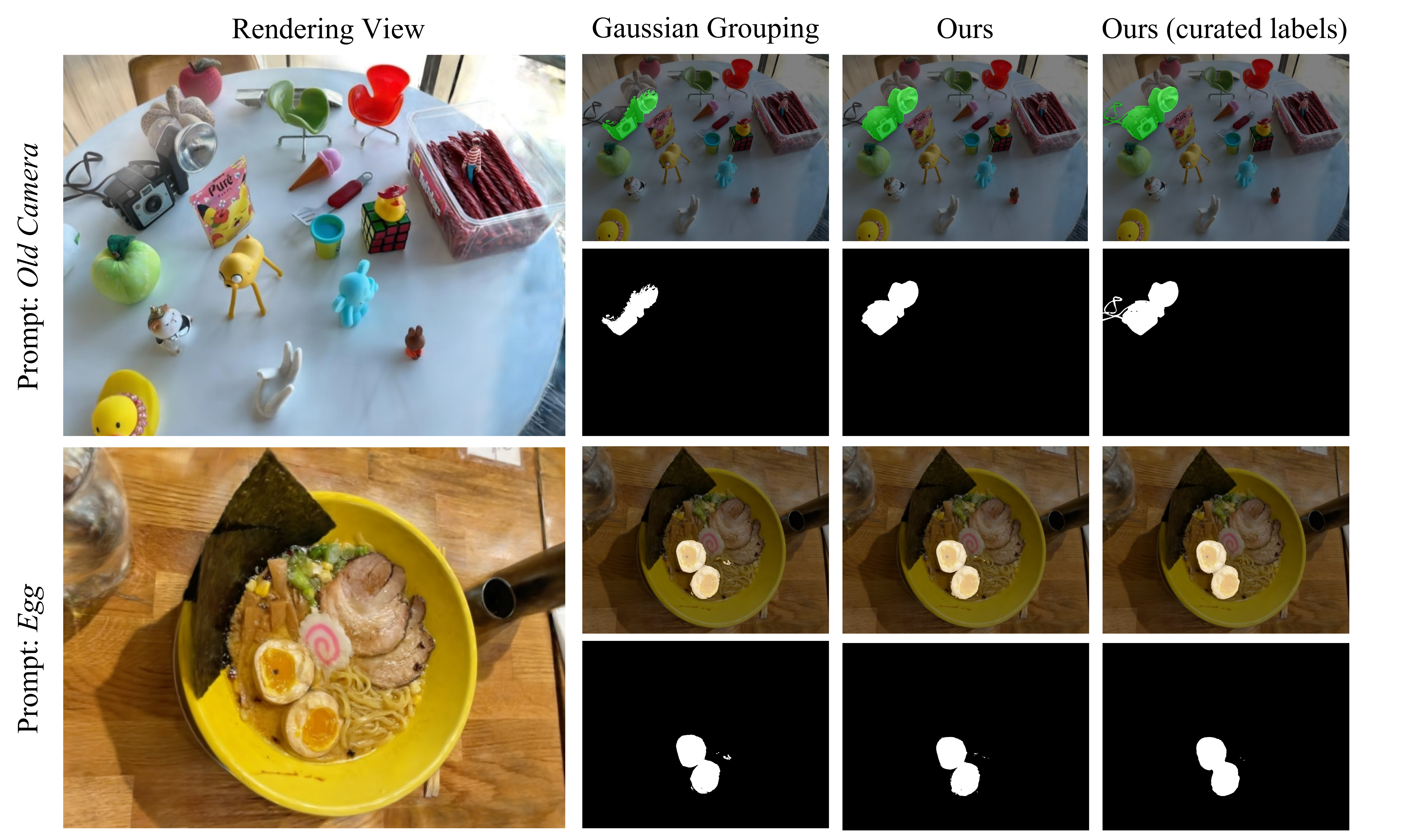}
  \caption{The visualization comparison results on the LERF-Mask dataset are as follows: For each scene, the first column shows the 3D reconstruction rendering results. For different text prompts, we use Grounding DINO to select the corresponding object IDs for rendering. The second column displays the results of Gaussian Grouping, and the third column shows our results. Additionally, we manually select the corresponding object IDs to demonstrate that our rendering results are sufficiently accurate.}
   \label{fig:lerf-mask-result}
\end{figure*}

\paragraph{Implementation Details}
For each scene, we train for 30,000 iterations: the first 12,000 apply the original Gaussian densification, followed by 3,000 IGD iterations, and densification operations are halted after 15,000 iterations. We use global KNN in the first 12,000 iterations, switching to LA-KNN between 12,000 and 30,000 iterations. The classification neural network uses a 1x1 convolution kernel, with the Identity Encoding dimension set to 16, resulting in 16 input channels and 256 output channels (where pixel values from 0-255 indicate classes). Softmax is applied after the convolution output to calculate class probabilities. The Adam optimizer and an A100 GPU are used, with parameters set as $\alpha = 1$, $\beta = 2$, $K = 5$, and $M = 1000$.
\subsection{Experimental Results and Analysis}

\begin{table}[t]
\centering
\resizebox{1\linewidth}{!}{
\begin{tabular}{l|cccc}
\toprule
\multirow{2}{*}{Method} & \multicolumn{4}{c}{mIoU} \\ \cline{2-5}
 & {figurines} & {ramen} & {teatime} & {average} \\ \midrule
OmniSeg3D~\citep{ying2024omniseg3d} & 69.7 & 77.0 & 71.7 & 79.4 \\
Feature3DGS~\citep{zhou2024feature} & 70.4 & 65.9 & 60.6 & 65.6 \\
GARField~\citep{kim2024garfield} & 89.2 & 75.7 & 77.8 & 80.9 \\
ClickGS~\citep{choi2024click} & \textbf{93.2} & \textbf{90.9} & 83.2 & 89.1 \\
GradiSeg (ours) & 90.1 & 89.0 & \textbf{89.7} & \textbf{89.6} \\ 
\bottomrule
\end{tabular}}
\caption{Comparison of 3D multi-view segmentation on the LERF-Mask dataset. The accuracy of the model's rendering results is evaluated by manually selecting object IDs for verification.}
\vspace{-2mm}
\label{tab:lerf-id}
\end{table}

\paragraph{Comparative Analysis of Open Vocabulary Segmentation}

To demonstrate the superior performance of our method in 3D segmentation tasks, we conduct an open vocabulary segmentation comparison against various baseline models on the LERF-Mask dataset. For each scene,  We adopt the detections from Grounding DINO~\citep{liu2023grounding} to select mask IDs in a 3D scene like Gaussian Grouping~\citep{ye2023gaussian}, and assess the quality of semantic segmentation by calculating the mean Intersection over Union (mIoU) and mean Boundary Intersection over Union (mBIoU). As shown in Table\ref{tab:lerf}, our method clearly outperforms all baseline models, confirming its capability to achieve high-quality 3D segmentation results. Notably, compared to Gaussian Grouping, the two proposed modules significantly enhance the model's semantic segmentation capabilities. In addition, for the open vocabulary segmentation task, Grounding DINO may return incorrect masks for certain text prompts. Consequently, even though our model can render accurate masks, the final generated masks may be problematic as they are adjusted to align with the results from Grounding DINO.

\paragraph{Comparative Analysis of Multi-View Segmentation}

In addition to the open vocabulary segmentation comparison experiment, we also conduct a 3D multi-view segmentation comparison on the LERF-Mask dataset. Using the input Segmentation Mask, we manually select the model-rendered IDs for the corresponding 3D objects. We then compare the model's rendered output with the Ground Truth Mask to evaluate our method's performance in 3D multi-view segmentation. In Table~\ref{tab:lerf-id}, Our method outperforms all comparative baselines across various scenarios, with the exception of ClickGS, and demonstrates particularly strong results in the teatime scenario. Additionally, our approach surpasses the current state-of-the-art ClickGS in terms of average performance, underscoring the superior multi-view semantic segmentation capabilities of our method.
\begin{table}[t]
\centering
\resizebox{1\linewidth}{!}{
\begin{tabular}{l|c|c|c}
\toprule
{Method} & {PSNR} & {SSIM} & {LPIPS}  \\ 
\midrule
Gau-Group~\citep{ye2023gaussian} & 27.09 & 0.826 & 0.166\\
GradiSeg (ours) & 27.05   & 0.826 & 0.167\\
\bottomrule
\end{tabular}}
\caption{Reconstruction Comparison on  Mip-NeRF 360 dataset~\citep{barron2022mip}. The result demonstrates that our method maintains reconstruction quality without degradation. }
\vspace{-2mm}
\label{tab:reconstruction}
\end{table}
\paragraph{Visual Analysis of Semantic Segmentation} 
In Figure~\ref{fig:lerf-mask-result}, we conduct a qualitative visual analysis of segmentation on the LERF-Mask dataset, where our approach demonstrates superior semantic segmentation capabilities compared to Gaussian Grouping. For example, our method achieves improved object recognition with sharper boundary delineation for old camera. In the ramen scene, we obtain a more precise and cleaner segmentation of the egg, significantly reducing noise points and highlighting the strength of our boundary segmentation. Additionally, we further illustrate our segmentation quality by manually selecting IDs in our rendering results, demonstrating that our outcomes in open vocabulary tasks are influenced by errors in Grounding DINO’s generated masks. These results indicate that our method, by leveraging precise boundary understanding, not only achieves more refined boundary delineation but also effectively reduces noise points in segmentation, thereby improving the accuracy of semantic segmentation in complex scenes.

\paragraph{Comparative Analysis of Reconstruction Quality}
In addition to the comparative experiments on 3D segmentation, We also evaluate the reconstruction quality of 3D scenes by comparing our method with Gaussian Grouping, as shown in Table~\ref{tab:reconstruction}. Specifically, following~\citep{kerbl20233d}, we calculate the PSNR, SSIM, and LPIPS metrics for scene reconstruction across different scenes from the Mip-NeRF 360 dataset to assess reconstruction quality. The difference in scene reconstruction quality between our approach and the Gaussian Grouping is minimal, indicating that the introduction of two additional modules enhances semantic understanding without compromising reconstruction quality, despite modifying the Gaussian distribution of the scene.

\subsection{Ablation Study}
\paragraph{Ablation study on boundary enhanced modules}
In Table~\ref{tab:ablation}, we present ablation studies on IGD and LA-KNN to evaluate the impact of these modules on the overall performance of our method on the LERF-Mask dataset. The results demonstrate that the removal of either IGD or LA-KNN leads to a marked decrease in open vocabulary segmentation accuracy, underscoring the importance of these modules. Specifically, the complete model with both IGD and LA-KNN consistently outperforms the ablated versions, showing superior boundary precision and semantic consistency. This significant performance gap highlights the critical role of IGD in refining boundary regions and of LA-KNN in maintaining local feature coherence, both of which are essential to achieving high-quality semantic segmentation.
\begin{figure}[t]
  \centering
  \includegraphics[width=1\linewidth]{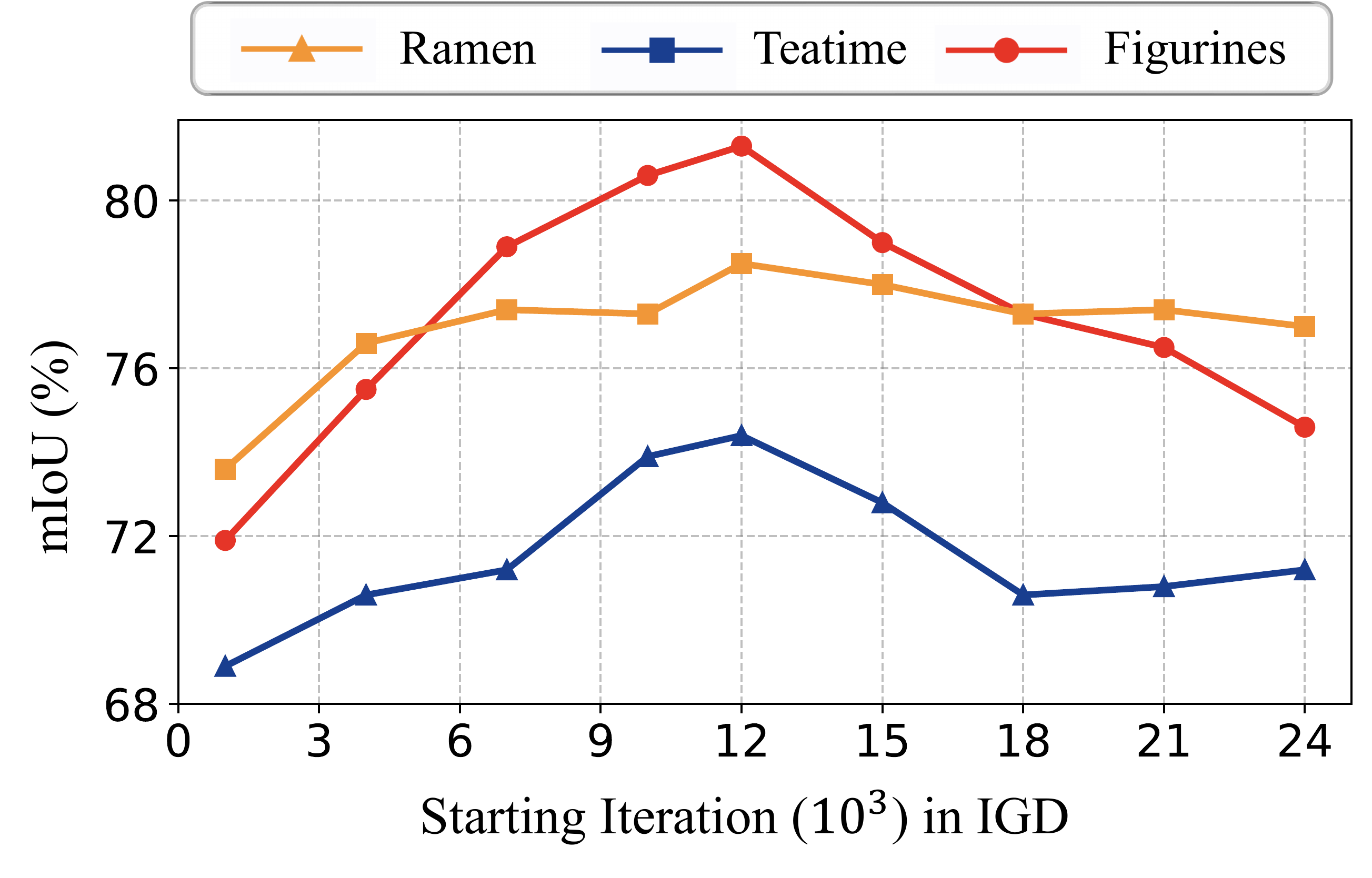}
  \vspace{-6mm}
  \caption{Ablation study on the impact of IGD starting iteration on LERF-Mask dataset. The result indicates that the starting iteration of IGD impacts the outcomes of semantic segmentation.}
   \label{fig:ablation_timing}
   \vspace{-4mm}
\end{figure}

\paragraph{Ablation study on IGD Start timing}
We conduct an ablation study to assess how the timing of IGD module activation impacts scene semantic segmentation. The Figure~\ref{fig:ablation_timing} shows that mIoU performance peaks around 12,000 iterations across three scenarios. Our analysis suggests that effective semantic segmentation depends on an initial geometric foundation. Once the scene reaches a certain reconstruction level, object boundaries are well-defined, allowing precise Gaussian adjustments. Activating IGD too late prevents the training process from converging to its optimal state, making it difficult to achieve the highest segmentation quality. This highlights that scene reconstruction and semantic understanding are not independent task.

\begin{figure}[t]
  \centering
  \includegraphics[width=1\linewidth]{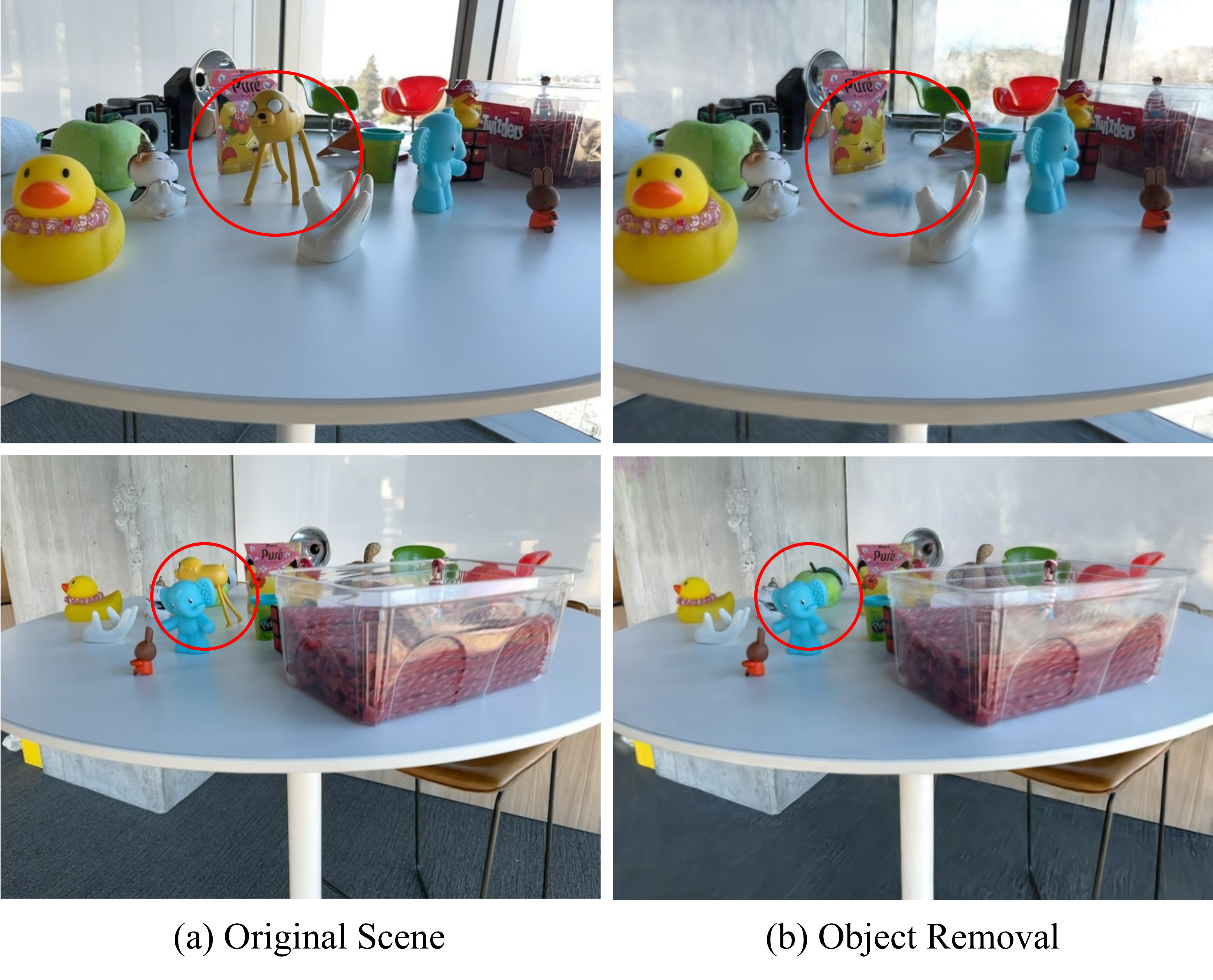}
  \vspace{-6mm}
  \caption{Besides semantic segmentation, our method also effectively supports scene editing tasks, such as object removal. By removing the object highlighted in the red circle from the original scene (column a), we obtain the object removal (column b).}
   \label{fig:removal}
\end{figure}
\begin{table}[t]
\centering
\resizebox{1\linewidth}{!}{
\begin{tabular}{l|cc|cc|cc}
\toprule
\multirow{2}{*}{Method} & \multicolumn{2}{c|}{figurines} & \multicolumn{2}{c|}{ramen} & \multicolumn{2}{c}{teatime}  \\ 
& mIoU & mBIoU & mIoU & mBIoU & mIoU & mBIoU    \\ \midrule 
\textbf{w/o IGD \& LA-KNN} &  69.7 & 67.9 & 77.0 & 68.7 & 71.7 & 66.1  \\
\textbf{w/o IGD}  &72.3 & 69.5 & 77.7  & 69.5 & 73.3 & 68.6 \\
\textbf{w/o LA-KNN}  &79.3 & 76.4 & 77.4  & 69.0 & 73.7 & 69.5 \\
\textbf{GradiSeg (ours)} & \textbf{81.3} & \textbf{78.1} & \textbf{78.5} & \textbf{72.9} & \textbf{74.4} & \textbf{70.6} \\ 
\bottomrule
\end{tabular}}
\caption{Ablation Studies on modules IGD and LA-KNN using LERF-Mask dataset.}
\vspace{-2mm}
\label{tab:ablation}
\end{table}
\subsection{Scene Editing}
Leveraging the decoupled segmentation representation enabled by Identity Encoding, along with our boundary-enhancement modules, our method more effectively supports downstream tasks such as scene editing, including object removal. By identifying all Gaussians in the 3D space that correspond to the particular Identity Encoding of an object and removing them as a group, the remaining Gaussians can be rendered to achieve object removal, as shown in the Figure~\ref{fig:removal}. Similarly, for other editing tasks, they are executed by manipulating the corresponding Gaussians.

%% file: sec/5_con.tex
\section{Conclusion}
\label{sec:con}
In this work, we propose GradiSeg, a novel framework for 3D semantic segmentation that effectively addresses issues of imprecise object boundaries in segmentation tasks. By integrating two boundary enhancement modules: Identity Gradient Guided Densification and Local Adaptive K-Nearest Neighbors, Our method adaptively refines Gaussians near boundaries while establishing Identity Encoding consistency within local 3D spaces, effectively preventing erroneous Gaussian feature propagation. These innovations enhance overall segmentation quality and mitigate boundary segmentation ambiguity. Comprehensive experiments on the LERF-Mask and MipNeRF 360 datasets demonstrate the superior performance of our GradiSeg.

%% file: sec/X_suppl.tex
\clearpage
\setcounter{page}{1}
\maketitlesupplementary
In this supplementary material, we provide more specific details of our method. In Section~\ref{sec:morimp}, we present more experimental details. In Section~\ref{sec:morres}, we provide additional qualitative results on LERF-Mask and Mip-NeRF 360 datasets. In Section~\ref{sec:dis}, we present analyses and discussions to clarify any confusing or unclear part of our method.

\section{More Implementation Details}
\label{sec:morimp}
\paragraph{More Implementation Details}
Following ~\citep{ye2023gaussian}, given a series of RGB images with associated poses, we leverage SAM~\citep{kirillov2023segment} to produce the corresponding segmentation masks, capitalizing on its outstanding segmentation capabilities. However, the 2D semantic masks generated by SAM from different viewpoints often have inconsistent mask IDs for the same object. To address this, we utilize DEVA~\citep{cheng2023trackingdecoupledvideosegmentation}, treating the images from various viewpoints as consecutive video frames and propagating the masks across views to ensure consistency.

During training, each scene is trained for 30,000 iterations, simultaneously optimizing the scene reconstruction loss and the semantic segmentation task. Gaussian densification and global KNN losses are applied during the first 12,000 iterations. The IGD module is activated between 12,000 and 15,000 iterations, followed by the application of the LA-KNN module from 15,000 to 30,000 iterations.

The 3D Identity Encoding is represented as a 16-dimensional vector with a total size of $N*1*16$, where 
$N$ denotes the number of Gaussians in the spatial domain. And the Identity Encoding gradient monitor is introduced with a size of $N*1$ to record the accumulated gradients of the identity encoding. After rendering the Identity Encoding, a feature map of size $16*H*W$ is obtained and subsequently fed into a classification network. For the classification network used during training, a $1*1$ convolution kernel is employed to adjust the dimensionality. The input dimension is set to 16, and the output dimension is set to 256. This design aligns with the semantic mask, where ID values are mapped to pixel values ranging from 0 to 255. Softmax is applied after the convolution output to calculate class probabilities. 

The entire training process is conducted using the Adam optimizer on an A100 GPU. The detailed parameter settings are as follows: $\alpha=1$, $\beta=2$, with the number $N$ of local neighbors set to 5 for both global KNN and LA-KNN. Additionally, the number $M$ of target Gaussians sampled for calculating the KL divergence is set to 1000.
\paragraph{More experiment Details}
\begin{table}[t]
\centering
\resizebox{1\linewidth}{!}{
\begin{tabular}{l|p{6cm}}
\toprule
Scene & Text Prompts  \\ 
\midrule
Figurines & green apple, porcelain hand, rubber duck with red hat, green toy chair, red apple, red toy chair, old camera \\ \hline
Ramen & chopsticks, pork belly, egg, wavy noodles in bowl, glass of water, yellow bowl \\ \hline
Teatime & apple, cookies on a plate, sheep, tea in a glass, bag of cookies, paper napkin, spoon handle, coffee mug, plate, stuffed bear \\
\bottomrule
\end{tabular}}
\caption{Text prompts during segmentation experiments on LERF-Mask dataset~\citep{ye2023gaussian}. }
\vspace{-2mm}
\label{tab:text}
\end{table}

 In the open-vocabulary semantic task, our evaluation method strictly adheres to that of Gaussian Grouping, utilizing the LERF-Mask dataset~\citep{ye2023gaussian}, which comprises three distinct scenes and 23 unique text query prompts, as shown in Table~\ref{tab:text}. Similarly, we employ Grounding DINO~\citep{liu2023grounding} to generate masks corresponding to the text prompts in order to match the id values of our rendered results. However, due to occasional inaccuracies in Grounding DINO's outputs, incorrect id values may be selected during the matching process, adversely affecting the final rendering results. To address this issue, we additionally manually select the object id values corresponding to the text query prompts based on the input results generated by DEVA and designed a multi-view semantic segmentation comparative experiment to verify the correctness of the model's rendering.
\section{More Results}
\label{sec:morres}
\begin{figure*}[t]
  \centering
  \includegraphics[width=1\linewidth]{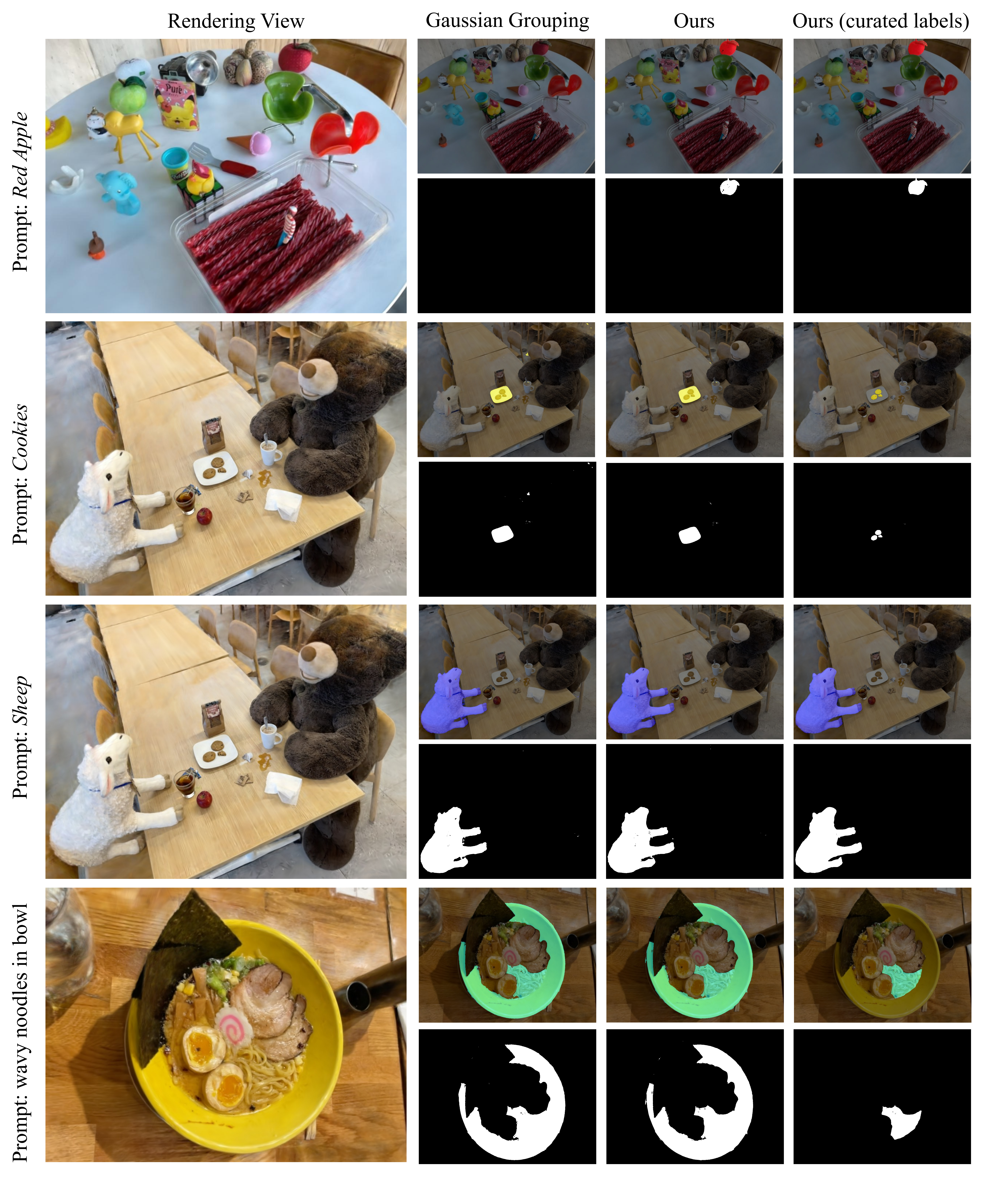}
  \caption{More visualization comparison results on the LERF-Mask dataset are as follows: For each scene, the first column shows the 3D reconstruction rendering results. The second column displays the results of Gaussian Grouping, and the third column shows our results. Additionally, we manually select the corresponding object IDs to demonstrate that our rendering results are sufficiently accurate.}
   \label{fig:sup-result}
\end{figure*}

\begin{figure*}[t]
  \centering
  \includegraphics[width=1\linewidth]{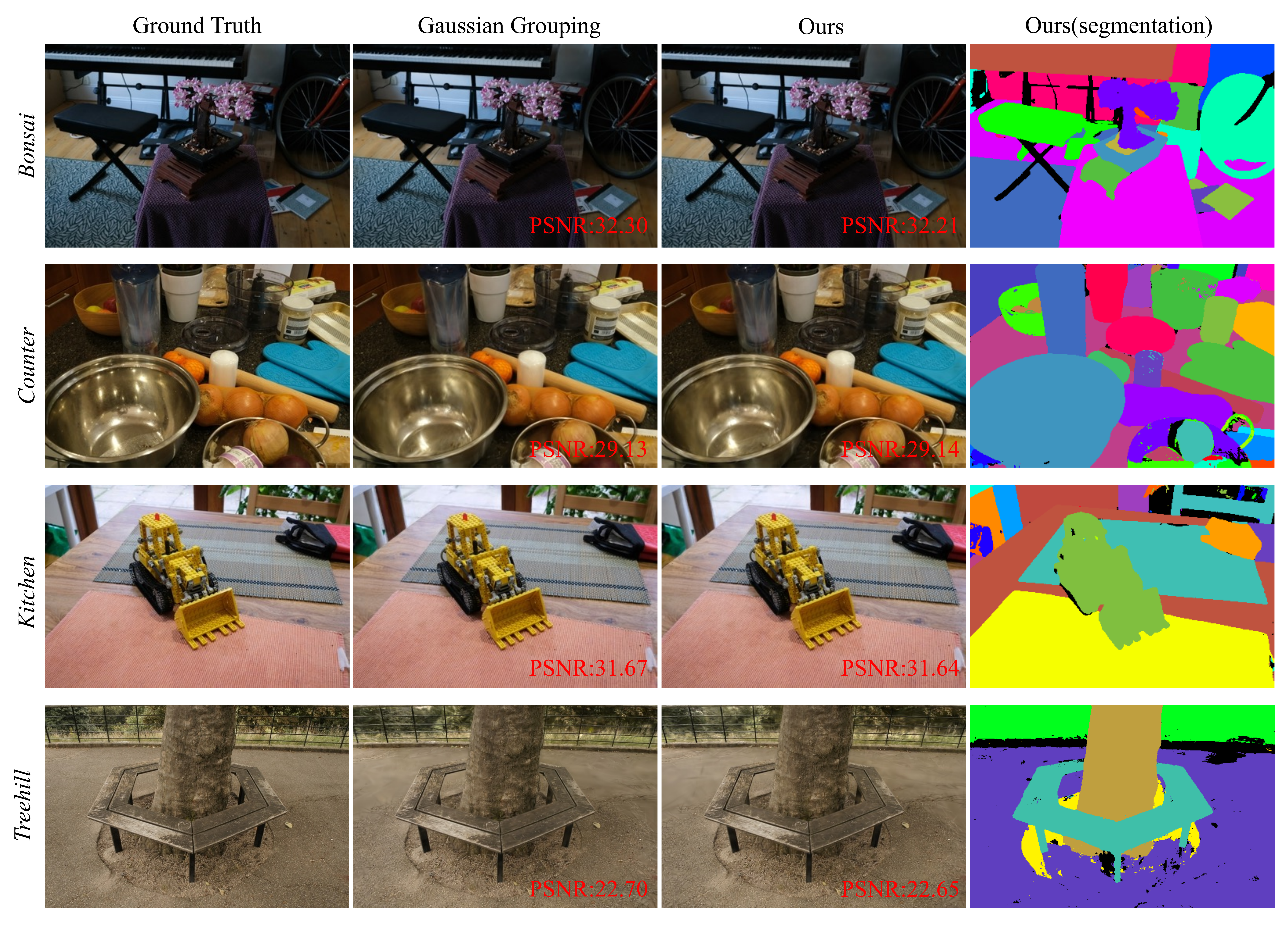}
  \caption{More visualization comparison results on the Mip-NeRF 360 dataset are as follows: For each scene, the first column shows the ground truth. The second column displays the reconstruction of Gaussian Grouping, and the third column shows our reconstruction. And the last column shows the segmentation of scene using our methond. Compared to Gaussian Grouping, our method enhances semantic segmentation without compromising the quality of scene reconstruction.}
   \label{fig:sup-re}
\end{figure*}

\begin{figure*}[t]
  \centering
  \includegraphics[width=1\linewidth]{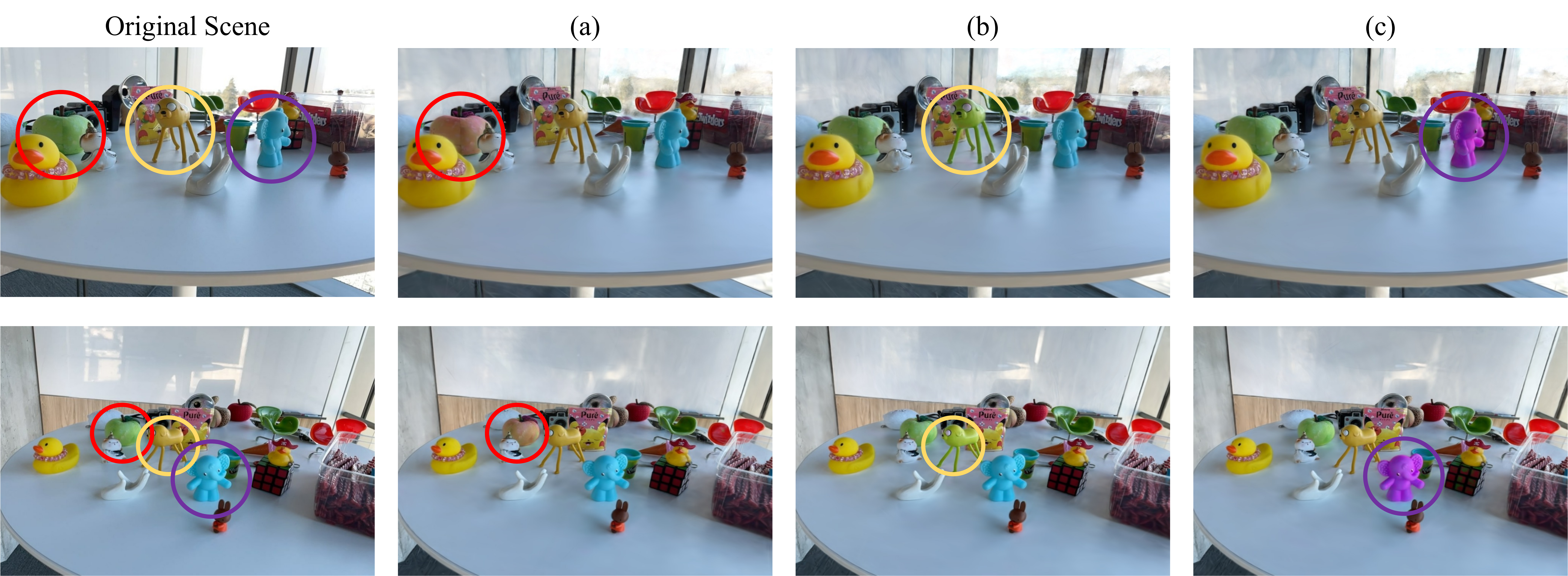}
  \caption{The visualization results demonstrate changes in group colors on the LERF-Mask dataset. For the original scene containing three objects, we implemented color modifications, with the results presented as (a), (b), and (c).}
   \label{fig:color}
\end{figure*}

The Figure~\ref{fig:sup-result} shows more visualization comparison results on the LERF-Mask dataset. Additionally, the Figure~\ref{fig:sup-re} shows more visualization comparison results on the Mip-NeRF 360 dataset~\citep{barron2022mip}. Compared to Gaussian Grouping~\citep{ye2023gaussian}, our method enhances semantic segmentation without compromising the quality of scene reconstruction. In the Figure~\ref{fig:color}, we additionally demonstrate a scene editing task by altering the colors of objects within the scene, showcasing that specific downstream editing tasks can be achieved by manipulating the Gaussians of a particular group.

\section{Analyses and Discussions}
\label{sec:dis}
\textbf{Q1: Why is a splitting operation performed in the IGD module?}
\par
\noindent\textbf{A1:}
For Gaussians exhibiting anomalous Identity Encoding gradients, the majority are concentrated near object boundaries, which often results in rendering inaccuracies for identity encodings. To mitigate this issue, we propose splitting such Gaussians into two smaller sub-Gaussians, strategically distributing them on either side of the object boundary. This approach enables the sub-Gaussians to independently capture the Identity Encodings of distinct objects, effectively resolving optimization conflicts. By replacing the original large Gaussian with the sub-Gaussians for rendering, this method ensures greater precision.
\par
\noindent\textbf{Q2:Why is the IGD module activated only after the completion of the original Gaussian densification?}
\par
\noindent\textbf{A2:}
The IGD module is initiated after the densification of the original Gaussians to ensure that the foundation of the Gaussian distribution is well-established before addressing identity encoding gradient anomalies. Densification consolidates the Gaussians, enhancing their spatial consistency and reducing redundancy, which provides a robust baseline for further refinement.

If the IGD module were activated prematurely, it might encounter unstable distributions or incomplete representations, leading to suboptimal splitting decisions or exacerbating gradient conflicts. By deferring its activation, the IGD module can operate on a more stable and well-defined Gaussian set, effectively splitting and redistributing Gaussians near object boundaries to optimize identity encoding, as shown in Figure~\ref{fig:ablation_timing} This sequential process ensures both the structural integrity of the distribution and the precision of identity encoding, leading to improved rendering outcomes.
\par
\noindent\textbf{Q3:Why is LA-KNN more effective than global KNN for handling boundary segmentation?}
\par
\noindent\textbf{A3:}
Both global KNN and LA-KNN facilitate feature propagation for identity encoding in 3D space. However, for Gaussians located near object boundaries, enforcing consistency with the K-nearest neighbors' features may lead to incorrect alignment with features from another object across the boundary. In contrast, LA-KNN computes the directional relationship of neighbors, enabling Gaussians to select the correct local neighbors. This directional awareness helps prevent feature propagation errors, ensuring more accurate identity encoding near boundaries, as shown in Figure~\ref{fig:la-knn}.
\par
\noindent\textbf{Q4:Why is global KNN applied first before activating LA-KNN?}
\par
\noindent\textbf{A4:}
The use of global KNN as an initial step before activating LA-KNN serves to establish a baseline feature propagation across the entire 3D space. This global operation ensures that Gaussians are broadly aligned with their nearest neighbors, promoting general feature consistency and reducing noise throughout the dataset.

However, global KNN lacks the capability to differentiate between local contexts, particularly near object boundaries, where it may propagate features across unrelated regions. By deferring the activation of LA-KNN, we leverage its ability to compute directional relationships among neighbors and refine the Gaussian associations in these critical areas. This sequential approach—first applying global KNN for overall consistency, followed by LA-KNN for localized boundary refinement—balances computational efficiency and segmentation precision, ensuring robust identity encoding and feature propagation.
\section{Limitation}
\label{lim}
Our method leverages a target tracking mechanism to pre-generate multi-view consistent segmentation masks before model training. However, unsuccessful target tracking may result in the incorrect assignment of segmentation mask IDs, which hinders the model's ability to learn effectively and limits its segmentation performance to some extent.

Furthermore, for open-vocabulary semantic tasks, we utilize third-party models to generate masks corresponding to textual prompts, which are then matched with our rendering results. In this case, the open-vocabulary segmentation capability of our model is constrained by the performance of the third-party models.
\newpage

%% file: main.bbl
\begin{thebibliography}{41}
\providecommand{\natexlab}[1]{#1}
\providecommand{\url}[1]{\texttt{#1}}
\expandafter\ifx\csname urlstyle\endcsname\relax
  \providecommand{\doi}[1]{doi: #1}\else
  \providecommand{\doi}{doi: \begingroup \urlstyle{rm}\Url}\fi

\bibitem[Barron et~al.(2022)Barron, Mildenhall, Verbin, Srinivasan, and Hedman]{barron2022mip}
Jonathan~T Barron, Ben Mildenhall, Dor Verbin, Pratul~P Srinivasan, and Peter Hedman.
\newblock Mip-nerf 360: Unbounded anti-aliased neural radiance fields.
\newblock In \emph{Proceedings of the IEEE/CVF conference on computer vision and pattern recognition}, pages 5470--5479, 2022.

\bibitem[Cen et~al.(2023)Cen, Zhou, Fang, Shen, Xie, Jiang, Zhang, Tian, et~al.]{cen2023segment}
Jiazhong Cen, Zanwei Zhou, Jiemin Fang, Wei Shen, Lingxi Xie, Dongsheng Jiang, Xiaopeng Zhang, Qi Tian, et~al.
\newblock Segment anything in 3d with nerfs.
\newblock \emph{Advances in Neural Information Processing Systems}, 36:\penalty0 25971--25990, 2023.

\bibitem[Cen et~al.(2024)Cen, Fang, Yang, Xie, Zhang, Shen, and Tian]{cen2024segment3dgaussians}
Jiazhong Cen, Jiemin Fang, Chen Yang, Lingxi Xie, Xiaopeng Zhang, Wei Shen, and Qi Tian.
\newblock Segment any 3d gaussians, 2024.

\bibitem[Cheng et~al.(2023)Cheng, Oh, Price, Schwing, and Lee]{cheng2023trackingdecoupledvideosegmentation}
Ho~Kei Cheng, Seoung~Wug Oh, Brian Price, Alexander Schwing, and Joon-Young Lee.
\newblock Tracking anything with decoupled video segmentation, 2023.

\bibitem[Cheng et~al.(2024)Cheng, Long, Yang, Yao, Yin, Ma, Wang, and Chen]{cheng2024gaussianpro3dgaussiansplatting}
Kai Cheng, Xiaoxiao Long, Kaizhi Yang, Yao Yao, Wei Yin, Yuexin Ma, Wenping Wang, and Xuejin Chen.
\newblock Gaussianpro: 3d gaussian splatting with progressive propagation, 2024.

\bibitem[Choi et~al.(2024)Choi, Song, Kim, Kim, and Do]{choi2024click}
Seokhun Choi, Hyeonseop Song, Jaechul Kim, Taehyeong Kim, and Hoseok Do.
\newblock Click-gaussian: Interactive segmentation to any 3d gaussians.
\newblock \emph{arXiv preprint arXiv:2407.11793}, 2024.

\bibitem[Fridovich-Keil et~al.(2022)Fridovich-Keil, Yu, Tancik, Chen, Recht, and Kanazawa]{fridovich2022plenoxels}
Sara Fridovich-Keil, Alex Yu, Matthew Tancik, Qinhong Chen, Benjamin Recht, and Angjoo Kanazawa.
\newblock Plenoxels: Radiance fields without neural networks.
\newblock In \emph{Proceedings of the IEEE/CVF conference on computer vision and pattern recognition}, pages 5501--5510, 2022.

\bibitem[Girish et~al.(2024)Girish, Gupta, and Shrivastava]{girish2024eaglesefficientaccelerated3d}
Sharath Girish, Kamal Gupta, and Abhinav Shrivastava.
\newblock Eagles: Efficient accelerated 3d gaussians with lightweight encodings, 2024.

\bibitem[Hamdi et~al.(2024)Hamdi, Melas-Kyriazi, Mai, Qian, Liu, Vondrick, Ghanem, and Vedaldi]{hamdi2024gesgeneralizedexponentialsplatting}
Abdullah Hamdi, Luke Melas-Kyriazi, Jinjie Mai, Guocheng Qian, Ruoshi Liu, Carl Vondrick, Bernard Ghanem, and Andrea Vedaldi.
\newblock Ges: Generalized exponential splatting for efficient radiance field rendering, 2024.

\bibitem[Hu et~al.(2024)Hu, Gong, Sun, and Wang]{hu2024lowlatencypointcloud}
Yueyu Hu, Ran Gong, Qi Sun, and Yao Wang.
\newblock Low latency point cloud rendering with learned splatting, 2024.

\bibitem[Kerbl et~al.(2023)Kerbl, Kopanas, Leimk{\"u}hler, and Drettakis]{kerbl20233d}
Bernhard Kerbl, Georgios Kopanas, Thomas Leimk{\"u}hler, and George Drettakis.
\newblock 3d gaussian splatting for real-time radiance field rendering.
\newblock \emph{ACM Trans. Graph.}, 42\penalty0 (4):\penalty0 139--1, 2023.

\bibitem[Kerr et~al.(2023)Kerr, Kim, Goldberg, Kanazawa, and Tancik]{kerr2023lerflanguageembeddedradiance}
Justin Kerr, Chung~Min Kim, Ken Goldberg, Angjoo Kanazawa, and Matthew Tancik.
\newblock Lerf: Language embedded radiance fields, 2023.

\bibitem[Kim et~al.(2024)Kim, Wu, Kerr, Goldberg, Tancik, and Kanazawa]{kim2024garfield}
Chung~Min Kim, Mingxuan Wu, Justin Kerr, Ken Goldberg, Matthew Tancik, and Angjoo Kanazawa.
\newblock Garfield: Group anything with radiance fields.
\newblock In \emph{Proceedings of the IEEE/CVF Conference on Computer Vision and Pattern Recognition}, pages 21530--21539, 2024.

\bibitem[Kirillov et~al.(2023)Kirillov, Mintun, Ravi, Mao, Rolland, Gustafson, Xiao, Whitehead, Berg, Lo, et~al.]{kirillov2023segment}
Alexander Kirillov, Eric Mintun, Nikhila Ravi, Hanzi Mao, Chloe Rolland, Laura Gustafson, Tete Xiao, Spencer Whitehead, Alexander~C Berg, Wan-Yen Lo, et~al.
\newblock Segment anything.
\newblock In \emph{Proceedings of the IEEE/CVF International Conference on Computer Vision}, pages 4015--4026, 2023.

\bibitem[Kong et~al.(2023)Kong, Liu, Li, Chen, Zhang, Ren, Pan, Chen, and Liu]{Kong_2023_ICCV}
Lingdong Kong, Youquan Liu, Xin Li, Runnan Chen, Wenwei Zhang, Jiawei Ren, Liang Pan, Kai Chen, and Ziwei Liu.
\newblock Robo3d: Towards robust and reliable 3d perception against corruptions.
\newblock In \emph{Proceedings of the IEEE/CVF International Conference on Computer Vision (ICCV)}, pages 19994--20006, 2023.

\bibitem[Lin et~al.(2024{\natexlab{a}})Lin, Li, Tang, Liu, Liu, Liu, Lu, Wu, Xu, Yan, and Yang]{lin2024vastgaussianvast3dgaussians}
Jiaqi Lin, Zhihao Li, Xiao Tang, Jianzhuang Liu, Shiyong Liu, Jiayue Liu, Yangdi Lu, Xiaofei Wu, Songcen Xu, Youliang Yan, and Wenming Yang.
\newblock Vastgaussian: Vast 3d gaussians for large scene reconstruction, 2024{\natexlab{a}}.

\bibitem[Lin et~al.(2024{\natexlab{b}})Lin, Feng, and Zhu]{lin2024rtgsenablingrealtimegaussian}
Weikai Lin, Yu Feng, and Yuhao Zhu.
\newblock Rtgs: Enabling real-time gaussian splatting on mobile devices using efficiency-guided pruning and foveated rendering, 2024{\natexlab{b}}.

\bibitem[Lin et~al.(2023)Lin, Dai, Zhu, and Yao]{lin2023gaussianflow4dreconstructiondynamic}
Youtian Lin, Zuozhuo Dai, Siyu Zhu, and Yao Yao.
\newblock Gaussian-flow: 4d reconstruction with dynamic 3d gaussian particle, 2023.

\bibitem[Liu et~al.(2020)Liu, Gu, Zaw~Lin, Chua, and Theobalt]{liu2020neural}
Lingjie Liu, Jiatao Gu, Kyaw Zaw~Lin, Tat-Seng Chua, and Christian Theobalt.
\newblock Neural sparse voxel fields.
\newblock \emph{Advances in Neural Information Processing Systems}, 33:\penalty0 15651--15663, 2020.

\bibitem[Liu et~al.(2023)Liu, Zeng, Ren, Li, Zhang, Yang, Jiang, Li, Yang, Su, et~al.]{liu2023grounding}
Shilong Liu, Zhaoyang Zeng, Tianhe Ren, Feng Li, Hao Zhang, Jie Yang, Qing Jiang, Chunyuan Li, Jianwei Yang, Hang Su, et~al.
\newblock Grounding dino: Marrying dino with grounded pre-training for open-set object detection.
\newblock \emph{arXiv preprint arXiv:2303.05499}, 2023.

\bibitem[Liu et~al.(2024)Liu, Guan, Luo, Fan, Wang, Peng, and Zhang]{liu2024citygaussianrealtimehighqualitylargescale}
Yang Liu, He Guan, Chuanchen Luo, Lue Fan, Naiyan Wang, Junran Peng, and Zhaoxiang Zhang.
\newblock Citygaussian: Real-time high-quality large-scale scene rendering with gaussians, 2024.

\bibitem[Lu et~al.(2023)Lu, Yu, Xu, Xiangli, Wang, Lin, and Dai]{lu2023scaffoldgsstructured3dgaussians}
Tao Lu, Mulin Yu, Linning Xu, Yuanbo Xiangli, Limin Wang, Dahua Lin, and Bo Dai.
\newblock Scaffold-gs: Structured 3d gaussians for view-adaptive rendering, 2023.

\bibitem[Luiten et~al.(2023)Luiten, Kopanas, Leibe, and Ramanan]{luiten2023dynamic}
Jonathon Luiten, Georgios Kopanas, Bastian Leibe, and Deva Ramanan.
\newblock Dynamic 3d gaussians: Tracking by persistent dynamic view synthesis.
\newblock \emph{arXiv preprint arXiv:2308.09713}, 2023.

\bibitem[Mildenhall et~al.(2021)Mildenhall, Srinivasan, Tancik, Barron, Ramamoorthi, and Ng]{mildenhall2021nerf}
Ben Mildenhall, Pratul~P Srinivasan, Matthew Tancik, Jonathan~T Barron, Ravi Ramamoorthi, and Ren Ng.
\newblock Nerf: Representing scenes as neural radiance fields for view synthesis.
\newblock \emph{Communications of the ACM}, 65\penalty0 (1):\penalty0 99--106, 2021.

\bibitem[M{\"u}ller et~al.(2022)M{\"u}ller, Evans, Schied, and Keller]{muller2022instant}
Thomas M{\"u}ller, Alex Evans, Christoph Schied, and Alexander Keller.
\newblock Instant neural graphics primitives with a multiresolution hash encoding.
\newblock \emph{ACM transactions on graphics (TOG)}, 41\penalty0 (4):\penalty0 1--15, 2022.

\bibitem[Qin et~al.(2024)Qin, Li, Zhou, Wang, and Pfister]{qin2024langsplat3dlanguagegaussian}
Minghan Qin, Wanhua Li, Jiawei Zhou, Haoqian Wang, and Hanspeter Pfister.
\newblock Langsplat: 3d language gaussian splatting, 2024.

\bibitem[Schieber et~al.(2024)Schieber, Young, Langlotz, Zollmann, and Roth]{schieber2024semanticscontrolledgaussiansplattingoutdoor}
Hannah Schieber, Jacob Young, Tobias Langlotz, Stefanie Zollmann, and Daniel Roth.
\newblock Semantics-controlled gaussian splatting for outdoor scene reconstruction and rendering in virtual reality, 2024.

\bibitem[Wang et~al.(2024)Wang, Chen, Liao, Fan, and Zhang]{Wang_2024_CVPR}
Yuqi Wang, Yuntao Chen, Xingyu Liao, Lue Fan, and Zhaoxiang Zhang.
\newblock Panoocc: Unified occupancy representation for camera-based 3d panoptic segmentation.
\newblock In \emph{Proceedings of the IEEE/CVF Conference on Computer Vision and Pattern Recognition (CVPR)}, pages 17158--17168, 2024.

\bibitem[Wu et~al.(2024)Wu, Yi, Fang, Xie, Zhang, Wei, Liu, Tian, and Wang]{wu20244d}
Guanjun Wu, Taoran Yi, Jiemin Fang, Lingxi Xie, Xiaopeng Zhang, Wei Wei, Wenyu Liu, Qi Tian, and Xinggang Wang.
\newblock 4d gaussian splatting for real-time dynamic scene rendering.
\newblock In \emph{Proceedings of the IEEE/CVF Conference on Computer Vision and Pattern Recognition}, pages 20310--20320, 2024.

\bibitem[Yan et~al.(2024)Yan, Low, Chen, and Lee]{yan2024multiscale3dgaussiansplatting}
Zhiwen Yan, Weng~Fei Low, Yu Chen, and Gim~Hee Lee.
\newblock Multi-scale 3d gaussian splatting for anti-aliased rendering, 2024.

\bibitem[Yang and Liu(2021)]{yang2021tupper}
Zhiliu Yang and Chen Liu.
\newblock Tupper-map: Temporal and unified panoptic perception for 3d metric-semantic mapping.
\newblock In \emph{2021 IEEE/RSJ International Conference on Intelligent Robots and Systems (IROS)}, pages 1094--1101. IEEE, 2021.

\bibitem[Yang and Liu(2023)]{yang2023unified}
Zhiliu Yang and Chen Liu.
\newblock Unified perception and collaborative mapping for connected and autonomous vehicles.
\newblock \emph{IEEE Network}, 37\penalty0 (4):\penalty0 273--281, 2023.

\bibitem[Yang et~al.(2024)Yang, Gao, Zhou, Jiao, Zhang, and Jin]{yang2024deformable}
Ziyi Yang, Xinyu Gao, Wen Zhou, Shaohui Jiao, Yuqing Zhang, and Xiaogang Jin.
\newblock Deformable 3d gaussians for high-fidelity monocular dynamic scene reconstruction.
\newblock In \emph{Proceedings of the IEEE/CVF Conference on Computer Vision and Pattern Recognition}, pages 20331--20341, 2024.

\bibitem[Ye et~al.(2023)Ye, Danelljan, Yu, and Ke]{ye2023gaussian}
Mingqiao Ye, Martin Danelljan, Fisher Yu, and Lei Ke.
\newblock Gaussian grouping: Segment and edit anything in 3d scenes.
\newblock \emph{arXiv preprint arXiv:2312.00732}, 2023.

\bibitem[Ying et~al.(2024)Ying, Yin, Zhang, Wang, Yu, Huang, and Fang]{ying2024omniseg3d}
Haiyang Ying, Yixuan Yin, Jinzhi Zhang, Fan Wang, Tao Yu, Ruqi Huang, and Lu Fang.
\newblock Omniseg3d: Omniversal 3d segmentation via hierarchical contrastive learning.
\newblock In \emph{Proceedings of the IEEE/CVF Conference on Computer Vision and Pattern Recognition}, pages 20612--20622, 2024.

\bibitem[Yu et~al.(2023)Yu, Chen, Huang, Sattler, and Geiger]{yu2023mipsplattingaliasfree3dgaussian}
Zehao Yu, Anpei Chen, Binbin Huang, Torsten Sattler, and Andreas Geiger.
\newblock Mip-splatting: Alias-free 3d gaussian splatting, 2023.

\bibitem[Zarzar et~al.(2022)Zarzar, Rojas, Giancola, and Ghanem]{zarzar2022segnerf}
Jesus Zarzar, Sara Rojas, Silvio Giancola, and Bernard Ghanem.
\newblock Segnerf: 3d part segmentation with neural radiance fields.
\newblock \emph{arXiv preprint arXiv:2211.11215}, 2022.

\bibitem[Zhang et~al.(2024)Zhang, Yang, Zuo, Tong, Long, and Liu]{zhang2024garfieldreinforcedgaussianradiance}
Hanyue Zhang, Zhiliu Yang, Xinhe Zuo, Yuxin Tong, Ying Long, and Chen Liu.
\newblock Garfield++: Reinforced gaussian radiance fields for large-scale 3d scene reconstruction, 2024.

\bibitem[Zhang et~al.(2020)Zhang, Riegler, Snavely, and Koltun]{zhang2020nerf++}
Kai Zhang, Gernot Riegler, Noah Snavely, and Vladlen Koltun.
\newblock Nerf++: Analyzing and improving neural radiance fields.
\newblock \emph{arXiv preprint arXiv:2010.07492}, 2020.

\bibitem[Zhi et~al.(2021)Zhi, Laidlow, Leutenegger, and Davison]{zhi2021place}
Shuaifeng Zhi, Tristan Laidlow, Stefan Leutenegger, and Andrew~J Davison.
\newblock In-place scene labelling and understanding with implicit scene representation.
\newblock In \emph{Proceedings of the IEEE/CVF International Conference on Computer Vision}, pages 15838--15847, 2021.

\bibitem[Zhou et~al.(2024)Zhou, Chang, Jiang, Fan, Zhu, Xu, Chari, You, Wang, and Kadambi]{zhou2024feature}
Shijie Zhou, Haoran Chang, Sicheng Jiang, Zhiwen Fan, Zehao Zhu, Dejia Xu, Pradyumna Chari, Suya You, Zhangyang Wang, and Achuta Kadambi.
\newblock Feature 3dgs: Supercharging 3d gaussian splatting to enable distilled feature fields.
\newblock In \emph{Proceedings of the IEEE/CVF Conference on Computer Vision and Pattern Recognition}, pages 21676--21685, 2024.

\end{thebibliography}
